\documentclass{mva_style}
\usepackage{graphicx}
\usepackage{color}
\usepackage{bm}
\usepackage{arydshln}
\usepackage[fleqn]{amsmath}

\finalcopy %Uncomment this line for the Camera-Ready Manuscript

\begin{document}
\title{Domain Generalization of Pathological Image Segmentation by Patch-Level and WSI-Level Contrastive Learning}

% \thanks{This study is supported by Grant-in-Aid for Challenging Research (Exploratory) JP23K18509, Cross-ministerial Strategic Innovation Promotion Program (SIP) on ``Integrated Health Care System'' Grant Number JPJ012425, and JST ASPIRE Program Japan Grant Number JPMJAP2403.} 

\author{
  Yuki Shigeyasu\\
  Kyushu Univ.\\
  Fukuoka, Japan\\
  % {\tt e-mail address1}\\
  \and
  % \textcolor{red}{Leave this area blank for double-blind review!}\\
  Shota Harada\\
  Kyushu Univ.\\
  % Kyushu University\\
  Fukuoka, Japan\\
  % {\tt harada@ait.kyushu-u.ac.jp}\\
  \and
  Akihiko Yoshizawa\\
  Nara Medical Univ.\\
  % Nara Medical University\\
  Nara, Japan\\
  \and
  Kazuhiro Terada\\
  Kyoto Univ. Hosp.\\
  Kyoto, Japan\\
  \and
  Naoki Nakazima\\
  Kyoto Univ. Hosp.\\
  % Kyoto University Hospital\\
  Kyoto, Japan\\
  \and
  Mariyo Kurata\\
  Kyoto Univ. Hosp.\\
  % Kyoto University\\
  Kyoto, Japan\\
  \and
  Hiroyuki Abe\\
  The Univ. of Tokyo Hosp.\\
  % The University of Tokyo Hospital\\
  Tokyo, Japan\\
  \and
  Tetsuo Ushiku\\
  The Univ. of Tokyo Hosp.\\
  % The University of Tokyo Hospital\\
  Tokyo, Japan\\
  \and
  Ryoma Bise\\
  Kyushu Univ.\\
  Fukuoka, Japan\\
  {\tt bise@ait.kyushu-u.ac.jp}
}

\maketitle

\section*{\centering Abstract}
\textit{
In this paper, we address domain shifts in pathological images by focusing on shifts within whole slide images~(WSIs), such as patient characteristics and tissue thickness, rather than shifts between hospitals. Traditional approaches rely on multi-hospital data, but data collection challenges often make this impractical. Therefore, the proposed domain generalization method captures and leverages intra-hospital domain shifts by clustering WSI-level features from non-tumor regions and treating these clusters as domains. To mitigate domain shift, we apply contrastive learning to reduce feature gaps between WSI pairs from different clusters. The proposed method introduces a two-stage contrastive learning approach WSI-level and patch-level contrastive learning to minimize these gaps effectively.\footnote{The original paper has been accepted in MVA2025. ©2025 IEICE. Personal use of this material is permitted. Permission from IEICE must be obtained for all other uses.}
}

\section{Introduction}
\begin{figure*}[t]
    \centering
    \includegraphics[width=.8\linewidth]{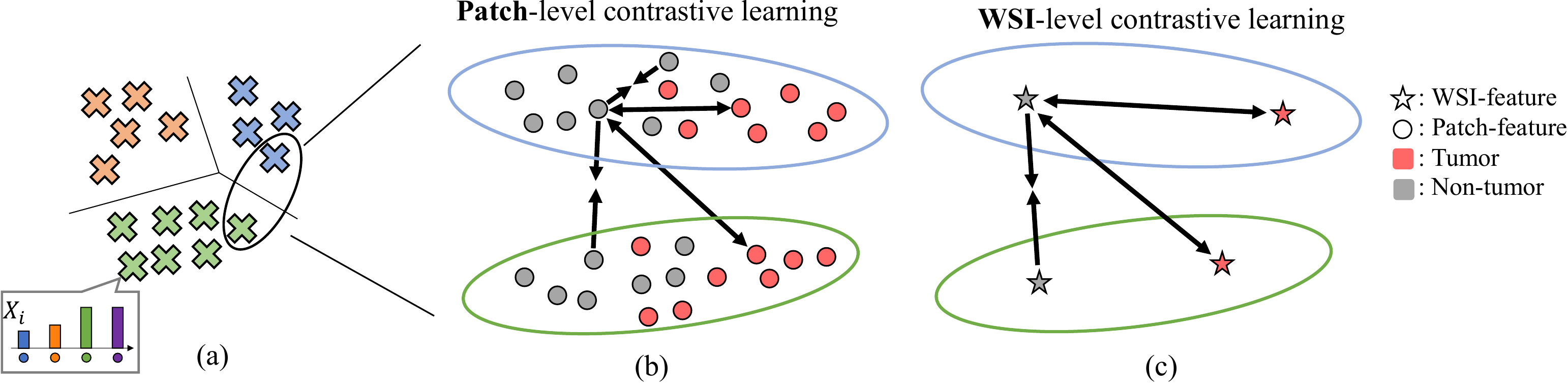}
    % \vspace{-0.5cm}
    \caption{Overview of contrastive learning. (a) WSI clustering result for pseudo domain construction, (b) Patch-level contrastive learning, (c) WSI-level contrastive learning with class-specific prototypes.}
    \label{fig:contrastive}
\end{figure*}
Domain shift occurs when training and test datasets have different distributions, leading to performance degradation. In recent years, with the advancement of machine learning, particularly deep learning, various challenges in the field of computer vision have been addressed using machine learning models. These methods are often designed assuming that the training and test datasets share the same distribution. However, this assumption may not hold in real-world scenarios. For instance, in pathology imaging, a model trained on data from one hospital may struggle with data from another hospital. Thus, domain shift remains a critical issue in practical applications \cite{muandet2013domain}.

To address domain shift problems, domain generalization has been studied. In traditional domain generalization methods, the model is trained to learn domain-independent feature representations using data labeled with multiple domains. In methods for pathology images, datasets collected from different hospitals typically exhibit domain shifts and are thus treated as separate domains. However, gathering data from multiple hospitals can be challenging due to issues such as privacy concerns and the need to comply with regulations like patient confidentiality. This setting corresponds to single-source domain generalization, where only one labeled domain is available and the model is expected to generalize to unknown domains without any additional supervision.

To avoid the challenges of data collection, we focus on the fact that domain shift may occur not only between different hospitals but also within a single hospital, due to various factors affecting whole slide images (WSIs), such as staining reagents, imaging devices, and tissue thickness or size. These factors are not always consistent within a single hospital, as the characteristics of cells can vary from patient to patient. Therefore, domain shift may occur even within a single hospital. Moreover, domain shifts do not occur within a single WSI, but rather between different WSIs. Within the same WSI, however, the manner in which the domain shift appears (i.e., how feature distributions shift) tends to follow a similar pattern.

In this study, we propose a single-source domain generalization method to address differences in image features caused by domain shifts in pathological images, using data collected from a single hospital. The key contributions are the joint learning of patch-level and WSI-level (patient-level) contrastive learning, as well as the creation of effective WSI and patch pairs, which address the challenges posed by the large size and complex tissue structures of WSIs. We evaluate the effectiveness of the proposed method in handling domain shifts caused by hospitals not used in training, based on experiments conducted with real clinical data.

\begin{figure*}[t]
    \centering
    \includegraphics[width=.8\linewidth]{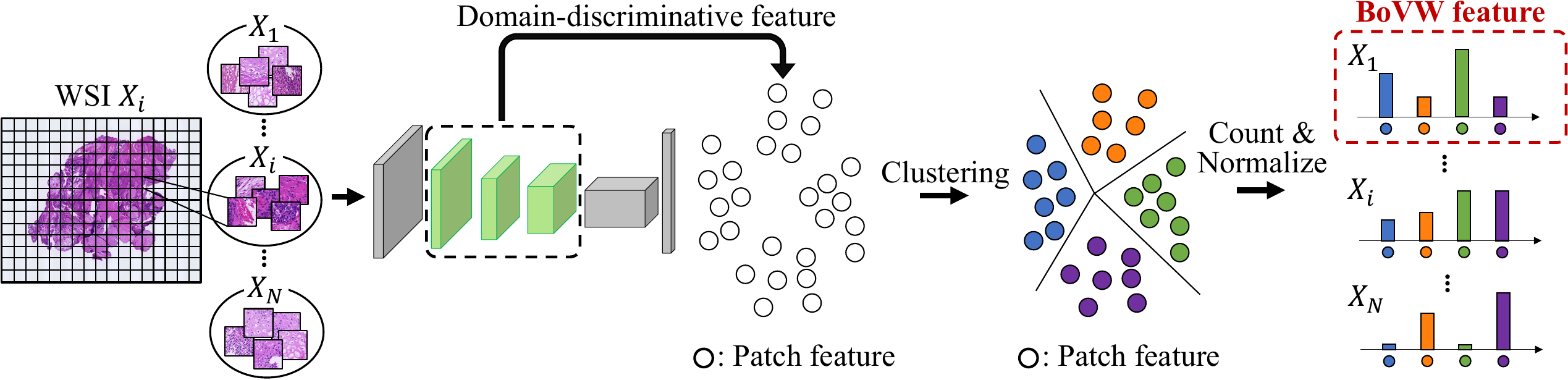}
    % \includegraphics[width=.8\linewidth]{figures/bovw_v1-crop.pdf}
    % \vspace{-0.8cm}
    \caption{Grouping of WSIs}
    \label{fig:wsi_feature}
\end{figure*}

\section{Related Work}
\noindent
\textbf{Domain generalization:}
Domain generalization learns domain-independent feature representations using only source data from multiple domains, aiming to improve classification on unknown target domains. Representative domain generalization methods include contrastive learning \cite{yoon2019generalizable}, adversarial learning \cite{li2018domain, akuzawa2020adversarial}, meta-learning \cite{li2018learning, zhao2021learning}, and data augmentation \cite{zhou2023semi, borlino2021rethinking}. However, these methods typically require multi-domain datasets, making them impractical for pathology image segmentation, where acquiring data from multiple hospitals is challenging.

\noindent
\textbf{Single-source domain generalization:}
Single-source domain generalization method improves classification performance on unknown target domains using a single domain for training. Additionally, since it typically ignores domain labels, it can also handle multiple source domains. Representative approaches include adversarial learning-based methods for data augmentation, which increase the diversity of the source domain \cite{fan2021adversarially, li2021progressive, wang2021learning}, methods that remove features strongly dependent on the domain \cite{huang2020self}, and methods that adapt to test data during testing \cite{liu2022single}. Unlike these methods, in this study, we propose a domain generalization approach that addresses domain shift even within the same hospital, which is traditionally considered a single domain. We aim to resolve this domain shift by performing contrastive learning using WSI features.

\noindent
\textbf{Contrastive learning:}
Contrastive learning trains models to bring similar data points closer and push dissimilar ones in the feature space. It is commonly used in self-supervised learning, where augmented data is treated as similar data, and all other data are considered dissimilar. Representative methods include SimCLR \cite{simclr}, which demonstrates improvements in accuracy through various data augmentation strategies, the use of a two-layer MLP, and increasing batch sizes. In supervised contrastive learning \cite{SupCon}, same-class data are treated as similar and data from different classes as dissimilar. Our method follows this approach, applying contrastive learning accordingly.

\section{Proposed Method}
Our method performs domain generalization through the joint learning of patch-level and WSI-level contrastive learning, as shown in Fig.~\ref{fig:contrastive}. To effectively create pairs of WSI features for contrastive learning, our method extracts the WSI features using the Bag of Visual Words (BoVW) approach, followed by clustering, as shown in Fig.~\ref{fig:wsi_feature}. Based on the clusters, WSI features from different groups are used to form negative pairs, and those from the same group are used to form positive pairs.

\subsection{Problem setting}
Let us assume that a set of $M$ source WSIs $\{\bm{X}_i, \bm{S}_i\}_{i=1}^{M}$ is given as the training data. Here, $\bm{X}$ represents the original WSI, and $\bm{S}$ is the ground truth segmentation mask. Since each WSI $\bm{X}_i$ is of very high resolution, directly inputting it into the model is difficult. Following existing studies \cite{tokunagaH2019, Li_2023_CVPR}, we handle the WSI segmentation task by dividing the WSI into patches, thereby converting the segmentation task into a classification task for patch images. Let the set of $L$ patch images from the source data be represented as $\mathcal{D} = \{\bm{x}_j, y_j, id\}_{j=1}^{L}$. Here, $\bm{x}_j \in \bm{X}_{id}$ denotes a patch image cut from the WSI, $id$ is the index of the WSI, and $y_j \in \{0, 1\}$ represents the class label, where 0 corresponds to non-tumor and 1 corresponds to tumor. In this task, given $\mathcal{D}$, the model is trained through domain generalization so that the trained model performs effectively on data from hospitals not used during training.

\subsection{Grouping WSIs using BoVW}
Domain shift in WSIs arises from factors like staining reagents, tissue thickness, and patient characteristics, affecting entire WSIs rather than individual cells. Thus, a common domain shift is assumed within each WSI. To address this, we group WSIs using BoVW-like strategy, treating each group as a pseudo-domain, and contrastive learning is then employed to reduce inter-domain differences. Clustering is performed using domain-discriminative features—independent of class-related characteristics—to facilitate effective domain generalization~\cite{MMLD}.

% Fig.~\ref{fig:wsi_feature} illustrates the method for grouping WSIs. When grouping WSIs, using global features might be influenced by factors such as the proportion of cancer types or cancerous regions. In contrast, BoVW better retains the distributional variance of local features, which is more informative for modeling staining-related domain shifts. For example, consider a case where one WSI contains 80\% tumor regions and another contains only 20\%. If we simply extract global WSI features from these images, the WSI features will differ due to the varying proportions of tumor regions, even if the patch-level features of tumor and non-tumor areas are similar.

Fig.~\ref{fig:wsi_feature} shows the WSI grouping method. Global features can be biased by cancer type ratios or tumor region proportions, while BoVW better preserves local feature distributions relevant to staining-related domain shifts. For instance, WSIs with different tumor region ratios may have dissimilar global features despite similar patch-level features.

To mitigate this, we focus on the non-tumor regions for feature extraction. We assume that factors affecting domain shift, such as staining reagents, imaging devices, and tissue thickness or size, are consistent within a WSI, meaning that the direction of shift across different patch images within a WSI is similar. Additionally, the appearance of non-tumor regions is more consistent than that of tumor regions. Therefore, we use non-tumor regions, which can better represent the domain shift of a WSI, for feature extraction.

% We obtain WSI features using the BoVW algorithm. Specifically, we first extract domain-discriminative features~\cite{MMLD} from ``non-tumor'' patches across all WSIs and cluster these features via K-means clustering. For the BoVW features of the $i$-th WSI, we count how many "non-tumor" patch features belong to each cluster, normalize the counts, and use them to form a BoVW feature vector for the $i$-th WSI, with the vector's dimension corresponding to the number of clusters. This BoVW WSI feature represents the distribution of ``non-tumor'' patch features within the WSI. We assume that if the distance between two WSI features is large, the domain shift is significant, and vice versa.
We extract WSI features using the BoVW algorithm. Domain-discriminative features~\cite{MMLD} from ``non-tumor'' patches are clustered via K-means. For each WSI, we count the number of patch features per cluster, normalize them, and use the result as a BoVW feature vector. This vector captures the distribution of ``non-tumor'' features. A larger distance between two such vectors indicates a greater domain shift.

To make pairs for contrastive learning, we perform clustering to the distribution of BoVW WSI features using K-means method. The resulting clusters represent the groups, and when performing contrastive learning, pairs are created between data from different groups. This approach is expected to create efficient pairs for learning.

\subsection{WSI-based contrastive learning}
% Pairs are created using the method described above, and contrastive learning is performed based on WSI features. When domain generalization is applied at the patch level, as in typical domain generalization methods, domain shift between patches is mitigated. However, since domain shift in WSIs is believed to occur at the WSI level rather than at the patch level, it is effective to address domain shift at the WSI level. Therefore, in this study, to resolve domain shift, we introduce contrastive learning not only between patches but also between WSIs. This approach ensures that domain shift is effectively addressed at both the local (patch) level and the global (WSI) level.
Using the above method, we construct pairs and perform contrastive learning based on WSI features. While typical domain generalization mitigates patch-level shifts, WSI-level shifts are more dominant in our setting. Therefore, we apply contrastive learning at both the patch and WSI levels to effectively address domain shift on both local and global level.

\noindent
\textbf{WSI-level contrastive learning:}
% We perform WSI-level contrastive learning to address domain shifts across entire WSIs. To effectively obtain domain-discriminative features, we bring distant WSI features (indicating large domain shifts) closer by randomly selecting feature pairs from different clusters using BoVW features. For representation learning, we use prototypes $\bm{c} \in \mathcal{C}$, which represent the mean feature of each WSI class (non-tumor and tumor), as shown in Fig.~\ref{fig:contrastive}(c). Contrastive learning treats same-class WSI features as positive pairs and different-class features as negative pairs, pulling the former closer and pushing the latter apart. Fig.\ref{fig:contrastive}(c) illustrates an example where the non-tumor feature in the upper WSI serves as the anchor.
To address domain shifts across WSIs, we apply WSI-level contrastive learning. Using BoVW features, we randomly sample feature pairs from different clusters—representing large domain shifts—and bring them closer. For representation learning, we use class prototypes $\bm{c} \in \mathcal{C}$, which are mean features for each class (non-tumor and tumor), as shown in Fig.~\ref{fig:contrastive}(c). Contrastive learning pulls same-class WSI features (positive pairs) closer and pushes different-class features (negative pairs) apart. Fig.~\ref{fig:contrastive}(c) shows an example where the non-tumor feature of the top WSI is used as the anchor.

Let the set of features for positive pairs in the source dataset $\mathcal{D}$ be denoted as $P_w(\bm{c}, \mathcal{C})$, and the set of features for negative pairs as $N_w(\bm{c}, \mathcal{C})$. The loss function for WSI-level contrastive learning between WSIs denoted as $L_w$ is as follows:
\footnotesize
\begin{equation}
L_w = \!\!\!
\sum_{\bm{c}_j\in\mathcal{C}}\!\frac{-1}{|P\left(\bm{c}_j,\mathcal{C}\right)|}
\!\!
\sum_{\substack{\bm{c}^+\in\\ P_w\left(\bm{c}_j,\mathcal{C}\right)}}
\!\!\!\!\!\
\log \frac{\displaystyle S\left(\bm{c}_j,\bm{c}^+\right)}{\displaystyle S\left(\bm{c}_j,\bm{c}^+\right)+
\!\!\!\!\!\!\!
\sum_{\substack{\bm{c}^-\in \\N_w\left(\bm{c}_j,\mathcal{C}\right)}}
\!\!\!\!\!
S\left(\bm{c}_j,\bm{c}^-\right)},
\end{equation}
\normalsize
where $\bm{c}_j$ is an anchor, $\bm{c}^+$, $\bm{c}^-$ are positive and negative samples.

This approach encourages patches belonging to the same WSI to move in the same direction, thus bringing the distributions of the same class closer together and pushing the distributions of different classes further apart.

\noindent
\textbf{Patch-level contrastive learning:}
To align the patch-level features between WSIs, patch-level contrastive learning is performed. Specifically, in patch-level contrastive learning, we select patch-level positive and negative pairs from the WSI-level pairs that are used for WSI-level contrastive learning. The anchor refers to the reference data, the positive example refers to data from the same class as the anchor, and the negative example refers to data from a different class than the anchor, as shown in Figure \ref{fig:contrastive} (b).

Let the patch-level feature vector $\bm{v}$ of anchor $\bm{x}$ be represented as $f(\bm{x}) = \bm{v} \in \mathcal{V}$, where $f$ is the feature extractor. Furthermore, let the set of features for positive pairs in the source dataset $\mathcal{D}$ be denoted by $P_p(\bm{v}, \mathcal{V})$, the set of features for negative pairs by $N_p(\bm{v}, \mathcal{V})$, and the similarity between feature vectors $\bm{v}_1$ and $\bm{v}_2$ be computed using a temperature parameter $\tau$. The loss function for patch-level contrastive learning between patches, denoted as $L_p$, is as follows:
\footnotesize
\begin{equation}
L_p = \!\!\!
\sum_{\bm{v}_j\in\mathcal{V}}\!\frac{-1}{|P\left(\bm{v}_j,\mathcal{V}\right)|}
\!\!
\sum_{\substack{\bm{v}^+\in\\ P_p\left(\bm{v}_j,\mathcal{V}\right)}}
\!\!\!\!\!\
\log \frac{\displaystyle S\left(\bm{v}_j,\bm{v}^+\right)}{\displaystyle S\left(\bm{v}_j,\bm{v}^+\right)+
\!\!\!\!\!\!\!
\sum_{\substack{\bm{v}^-\in \\N_p\left(\bm{v}_j,\mathcal{V}\right)}}
\!\!\!\!\!
S\left(\bm{v}_j,\bm{v}^-\right)}
\label{eq:contrastive}
\end{equation}
\normalsize
where $\bm{v}_j$ is an anchor, $\bm{v}^+$, $\bm{v}^-$ are patch-level positive and negative samples.

In addition to the contrastive learning loss functions for patches and WSIs, $L_p$ and $L_w$, the cross-entropy loss $L_c$ is also used to train the fully connected layer. The overall loss is defined as $L_w + L_p + L_c$.

\section{Experiments}
We evaluated the proposed method in a two-class (non-tumor, tumor) segmentation task using whole-slide images (WSIs) of cervix tissues. The source data, collected from Kyoto University Hospital, was used for training, while target data, collected from a different hospital (The University of Tokyo Hospital), was used for testing.

For the evaluation of segmentation performance, we used commonly employed metrics in segmentation tasks: precision, recall, F1 score, and macro-F1 score. Here, precision, recall, and F1 score treat the tumor class as positive, while macro-F1 represents the average F1 score, considering both the tumor and non-tumor classes as positive.

\noindent
\textbf{Dataset:}
% Cervical cancer slices were imaged using a virtual slide scanner at a maximum magnification of 40x after being stained with Hematoxylin and Eosin (H\&E). The source dataset includes 300 WSIs, and the target dataset includes 108 WSIs. In the experiment, 236 WSIs from the source data were used for training, 64 WSIs for validation, and all target WSIs for testing. After splitting the data, $256 \times 256$ patches were extracted at 40x magnification. Only patches whose ground-truth masks consisted entirely of a single class were used for training. The source data contained a total of 356,079 patches, with 28,371 patches belonging to the tumor class and 327,708 patches belonging to the non-tumor class, while the target dataset had 68,164 patches, with 6,071 patches from the tumor class and 62,093 patches from the non-tumor class.
Cervical cancer tissue slices were stained with Hematoxylin and Eosin (H\&E) and scanned using a virtual slide scanner at 40× magnification. The source dataset consists of 300 WSIs, and the target dataset includes 108 WSIs. For the experiments, 236 WSIs from the source were used for training, 64 for validation, and all target WSIs for testing. After splitting, $256 \times 256$ patches were extracted at 40× magnification. Only patches whose ground-truth masks were entirely composed of a single class were used for training as that class. The source dataset yielded 356,079 patches in total—28,371 tumor and 327,708 non-tumor—while the target dataset contained 68,164 patches—6,071 tumor and 62,093 non-tumor.

\noindent
\textbf{Implementation details:}
For feature extraction, we used a modified version of ResNet-18 \cite{resnet} pretrained on ImageNet, replacing the final layer with a two-layer MLP. Training was performed using SGD with a learning rate of $10^{-5}$ and a batch size of 128. From each of the two WSIs, 32 patches were randomly sampled for both tumor and non-tumor classes. If a WSI had fewer than 32 patches for a class, oversampling was applied.

\noindent
\textbf{Performance evaluation:}
We compared segmentation performance across four methods: 1) a baseline method that uses only the cross-entropy loss for training on the source data, 2) a method that leverages style features to create domain-discriminative representations and assigns domain labels based on latent domains of the images through clustering for domain generalization \cite{MMLD} (MMLD), 3) a method that extracts style features using mutual information and performs domain generalization \cite{L2D} (L2D), and 4) the proposed method (Ours). In both MMLD and the proposed method, K-means clustering was used to assign domain labels, and the value of K was varied as 2, 4, 6, 8, and 10. The value that yielded the highest performance on the validation set was used for comparison.

\begin{table}[t]
    \def\@captype{table}
      \makeatother
        \centering
        \caption{Comparison of segmentation performance and ablation study.}
        % \vspace{-0.2cm}
        \resizebox{\columnwidth}{!}{
        \begin{tabular}{c|cccc} \hline
        Method & precision & recall & F1 & macro-F1\\ \hline
        Baseline & 0.4255 & 0.7218 & 0.5354 & 0.7360\\
        w/ contrastive & 0.4457 & 0.6697 & 0.5344 & 0.7380\\
        MMLD \cite{MMLD} & 0.4215 & 0.7457 & 0.5385 & 0.7368\\
        L2D \cite{L2D} & \bf{0.5090} & 0.5586 & 0.5326 & 0.7427 \\
        Ours & 0.4827 & \bf{0.7521} & \bf{0.5880} & \bf{0.7675}\\\hline
        \end{tabular}
        }
        \label{tab:comparison}
\end{table}

Table \ref{tab:comparison} shows the performance results for each method. Existing domain generalization methods show improvements in accuracy on several metrics compared to the Baseline. The proposed method, which performs domain generalization using WSI features, outperforms the Baseline on all metrics and achieves the highest F1 and macro-F1 scores, further improving performance compared to other domain generalization methods.

Table \ref{tab:comparison} also presents the results of the Ablation study. Baseline+contrastive (w/ contrastive) applies contrastive learning by using cross-entropy loss on source data, treating same-class patches as positive and different classes as negative. Compared to the simple cross-entropy loss (Baseline), Baseline+contrastive shows limited improvement in performance. However, the proposed method (Ours) improves performance across all metrics. This suggests that the performance improvement of the proposed method is not solely attributed to the introduction of general contrastive learning.

\section{Conclusion}
% In this paper, we proposed a method for addressing domain shift in pathology images, which occurs due to differences in factors such as tissue slice thickness and staining solutions. We hypothesized that domain shift occurs at the whole-slide image (WSI) level and introduced a method to mitigate domain shift at this level. Our method utilizes contrastive learning for domain shift improvement. In addition to conventional contrastive learning at the patch level, we introduced contrastive learning of the average features of patches to shift the distribution of the entire WSI. Experiments show our method outperforms existing domain generalization techniques by reducing domain shift and achieves state-of-the-art segmentation performance under realistic domain shift conditions, even without access to multi-domain training data.
In this paper, we proposed a method to address domain shift in pathology images, caused by factors such as tissue thickness and staining variations. Assuming that domain shift occurs at the whole-slide image (WSI) level, we introduced a contrastive learning approach that operates not only at the patch level but also at the WSI level using average patch features. Our method effectively reduces domain shift and outperforms existing domain generalization techniques, achieving state-of-the-art segmentation performance under realistic domain shift conditions without requiring multi-domain training data.
\par
% \subsubsection*{Acknowledgement}
\noindent
{\bf Acknowledgement}
This study is supported by Grant-in-Aid for Challenging Research (Exploratory) JP23K18509, Cross-ministerial Strategic Innovation Promotion Program (SIP) on ``Integrated Health Care System'' Grant Number JPJ012425, and JST ASPIRE Program Japan Grant Number JPMJAP2403.

\bibliographystyle{ieeetran}
\bibliography{myrefs}

% Generated by IEEEtran.bst, version: 1.14 (2015/08/26)
\begin{thebibliography}{10}
\providecommand{\url}[1]{#1}
\csname url@samestyle\endcsname
\providecommand{\newblock}{\relax}
\providecommand{\bibinfo}[2]{#2}
\providecommand{\BIBentrySTDinterwordspacing}{\spaceskip=0pt\relax}
\providecommand{\BIBentryALTinterwordstretchfactor}{4}
\providecommand{\BIBentryALTinterwordspacing}{\spaceskip=\fontdimen2\font plus
\BIBentryALTinterwordstretchfactor\fontdimen3\font minus \fontdimen4\font\relax}
\providecommand{\BIBforeignlanguage}[2]{{%
\expandafter\ifx\csname l@#1\endcsname\relax
\typeout{** WARNING: IEEEtran.bst: No hyphenation pattern has been}%
\typeout{** loaded for the language `#1'. Using the pattern for}%
\typeout{** the default language instead.}%
\else
\language=\csname l@#1\endcsname
\fi
#2}}
\providecommand{\BIBdecl}{\relax}
\BIBdecl

\bibitem{muandet2013domain}
K.~Muandet \emph{et~al.}, ``Domain generalization via invariant feature representation,'' in \emph{ICML}, vol.~28, 2013, pp. 10--18.

\bibitem{yoon2019generalizable}
C.~Yoon \emph{et~al.}, ``Generalizable feature learning in the presence of data bias and domain class imbalance with application to skin lesion classification,'' in \emph{MICCAI}, 2019, pp. 365--373.

\bibitem{li2018domain}
H.~Li \emph{et~al.}, ``Domain generalization with adversarial feature learning,'' in \emph{CVPR}, 2018.

\bibitem{akuzawa2020adversarial}
K.~Akuzawa \emph{et~al.}, ``Adversarial invariant feature learning with accuracy constraint for domain generalization,'' in \emph{ECML PKDD}, 2020, pp. 315--331.

\bibitem{li2018learning}
D.~Li \emph{et~al.}, ``Learning to generalize: Meta-learning for domain generalization,'' \emph{AAAI}, vol.~32, no.~1, 2018.

\bibitem{zhao2021learning}
Y.~Zhao \emph{et~al.}, ``Learning to generalize unseen domains via memory-based multi-source meta-learning for person re-identification,'' in \emph{CVPR}, 2021, pp. 6277--6286.

\bibitem{zhou2023semi}
K.~Zhou \emph{et~al.}, ``Semi-supervised domain generalization with stochastic stylematch,'' \emph{Int. J. Comput. Vis.}, pp. 1--11, 2023.

\bibitem{borlino2021rethinking}
F.~Borlino \emph{et~al.}, ``Rethinking domain generalization baselines,'' in \emph{ICPR}, 2021, pp. 9227--9233.

\bibitem{fan2021adversarially}
X.~Fan \emph{et~al.}, ``Adversarially adaptive normalization for single domain generalization,'' in \emph{CVPR}, 2021, pp. 8208--8217.

\bibitem{li2021progressive}
L.~Li \emph{et~al.}, ``Progressive domain expansion network for single domain generalization,'' in \emph{CVPR}, 2021, pp. 224--233.

\bibitem{wang2021learning}
Z.~Wang \emph{et~al.}, ``Learning to diversify for single domain generalization,'' in \emph{ICCV}, 2021, pp. 834--843.

\bibitem{huang2020self}
Z.~Huang \emph{et~al.}, ``Self-challenging improves cross-domain generalization,'' in \emph{ECCV}, 2020, pp. 124--140.

\bibitem{liu2022single}
Q.~Liu \emph{et~al.}, ``Single-domain generalization in medical image segmentation via test-time adaptation from shape dictionary,'' \emph{AAAI}, vol.~36, no.~2, pp. 1756--1764, 2022.

\bibitem{simclr}
T.~Chen \emph{et~al.}, ``A simple framework for contrastive learning of visual representations,'' in \emph{ICML}, 2020.

\bibitem{SupCon}
P.~Khosla \emph{et~al.}, ``Supervised contrastive learning,'' \emph{NeurIPS}, 2020.

\bibitem{tokunagaH2019}
H.~Tokunaga \emph{et~al.}, ``Adaptive weighting multi-field-of-view cnn for semantic segmentation in pathology,'' in \emph{CVPR}, 2019.

\bibitem{Li_2023_CVPR}
H.~Li \emph{et~al.}, ``Task-specific fine-tuning via variational information bottleneck for weakly-supervised pathology whole slide image classification,'' in \emph{CVPR}, 2023, pp. 7454--7463.

\bibitem{MMLD}
T.~Matsuura and T.~Harada, ``Domain generalization using a mixture of multiple latent domains,'' \emph{AAAI}, vol.~34, no.~07, pp. 11\,749--11\,756, 2020.

\bibitem{resnet}
K.~He \emph{et~al.}, ``Deep residual learning for image recognition,'' in \emph{CVPR}, 2016.

\bibitem{L2D}
Z.~Wang \emph{et~al.}, ``Learning to diversify for single domain generalization,'' in \emph{ICCV}, 2021, pp. 834--843.

\end{thebibliography}
\end{document}